# Velocity Selection for High-Speed UGVs in Rough Unknown Terrains using Force Prediction

Graeme N. Wilson, Alejandro Ramirez-Serrano, Mahmoud Mustafa, and Krispin A. Davies

Department of Mechanical and Manufacturing Engineering, University of Calgary, Calgary, Alberta, Canada
{wilsongn, aramirez, mmmustaf, kdavies}@ucalgary.ca

**Abstract.** Enabling high speed navigation of Unmanned Ground Vehicles (UGVs) in unknown rough terrain where limited or no information is available in advance requires the assessment of terrain in front of the UGV. Attempts have been made to predict the forces the terrain exerts on the UGV for the purpose of determining the maximum allowable velocity for a given terrain. However, current methods produce overly aggressive velocity profiles which could damage the UGV. This paper presents three novel safer methods of force prediction that produce effective velocity profiles. Two models, Instantaneous Elevation Change Model (IECM) and Sinusoidal Base Excitation Model: using Excitation Force (SBEM:EF), predict the forces exerted by the terrain on the vehicle at the ground contact point, while another method, Sinusoidal Base Excitation Model: using Transmitted Force (SBEM:TF), predicts the forces transmitted to the vehicle frame by the suspension.

## 1 Introduction

Autonomous traversal of unknown rough terrains at high-speeds is a challenging endeavor for Unmanned Ground Vehicles (UGVs). Information about unknown terrain must be gathered online using either proprioceptive or exteroceptive sensors to allow UGVs to avoid obstacles, achieve navigation goals, and maintain forces transmitted by the terrain on the vehicle at safe levels.

Proprioceptive detection of vehicle vibrations during terrain traversal has been used to classify terrain using trained probabilistic neural networks (PNNs)[1], [2] as well as using support vector machine (SVM) classifiers [3]. The problem with these methods is that significant offline training is required, they are dependent on the speeds at which the vehicles are trained, and they produce misclassifications of terrain during traversal. To resolve the speed dependency problem the frequency response of the terrain was obtained from the acceleration data using a transfer function in [4], but offline training is still required and misclassification of terrain still occurs. In addition to the previous stated problems, proprioceptive approaches are reactive, which means that large changes in terrain roughness may be undetected until after the vehicle encounters them.

To detect upcoming terrain changes exteroceptive sensors can be used. The combination of vibration and vision based classification using SVM has been used to classify upcoming terrain characteristics [5], [6]. These methods predict upcoming terrain at the expense of offline training. Online training methods using this combined approach have also been investigated; however, these approaches still present the potential for terrain misclassification which would damage the UGV [7], [8].

To prevent misclassification of terrain, methods have been developed which use geometric information about the terrain from stereo cameras and laser scanners [9–11]. Using a stereo camera a danger value is computed in [9] using terrain roughness, slope, and step height. While this work is useful for path planning it does not consider velocity selection for the UGV. In contrast, the authors in [10] present a fuzzy logic approach which outputs target velocity based on roughness and slope inputs. This approach allows for velocity control of UGV based on upcoming terrain; however, it does not guarantee that forces acting on the UGV are kept in a safe range.

Addressing the problem of maintaining safe forces, the work presented by [11] computes a Roughness Index (RI) based on the elevation of the terrain detected by a laser scanner. RI value is used to compute the allowable velocity for traversing upcoming terrain based on the predicted forces that the terrain will exert on the vehicle. While this approach solves many of the issues previously presented, the methods used in [11] to calculate the force exerted by the terrain produce aggressive velocity estimates which may still result in UGV damage.

To avoid UGV damage this paper presents three novel safer methods of predicting the force exerted by unknown terrain on a UGV. These new methods include two techniques that predict the base excitation force exerted by the terrain, and one technique predicting the force transmitted to the vehicle frame. These techniques use the assumption that elevation data follows a normal distribution; they thus calculate the worst case maximum terrain elevation from the RI. The potential for resonance in the suspension is even accounted for in the transmitted force model. These methods are designed to produce fast and safe values of maximum allowable velocity for rough unknown terrains.

## 2    Roughness Index

Developing force prediction models for high-speed UGVs requires a measure of the traversability of upcoming terrain. As described in [11] the Roughness Index (RI) can be used to provide a quantitative measure of terrain roughness from a 3D point cloud. In this approach the RI was described as a number ranging from 0 to 1, where 0 was the roughest perceived terrain. The problem with the approach proposed in [11] is that in many cases, such as a simple sinusoidal terrain profile, the RI becomes negative before the maximum terrain elevation exceeds the ground clearance of the vehicle. Since negative values of RI are considered untraversable in [11], certain terrain are falsely considered untraversable. To solve this problem the RI proposed in [11] is redefined here as shown in Eq. (1) where $e$ represents the terrain elevation for each point in the 3D point cloud and $h$ represents the vehicle ground clearance.

$$RI = STD\left(\frac{e}{h}\right) \tag{1}$$

The redefined RI is a number which ranges from 0 (smoothest terrain) to $\infty$ (roughest possible terrain). The RI at which the terrain is considered untraversable is a value which is defined separately for each application as it depends on the abilities of the vehicle being used. A comparison of the RI defined in this paper and the RI defined in [11] is shown in Fig. 1. In this figure the RI values are shown as calculated on a range of sinusoidal terrains with frequencies of $2\pi$ and amplitudes ranging from 0 to $0.1m$. The ground clearance of the vehicle was set at $0.1m$. For other values of ground clearance results will be similar except the slopes of the lines will decrease.

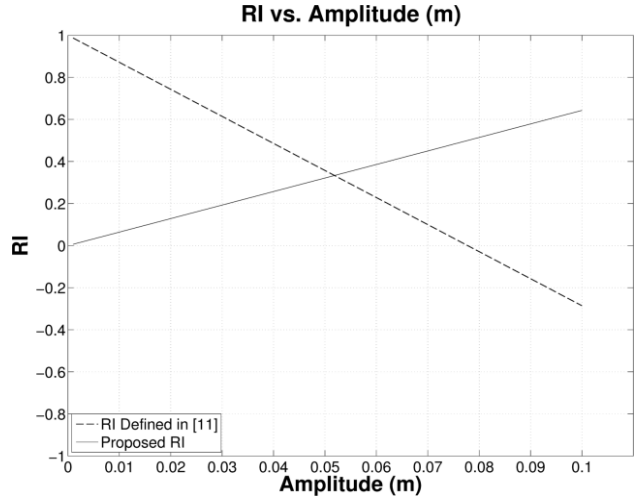

**Fig. 1.** Comparison of RI's – [11] vs. Proposed RI (Eq. (1))

From Fig. 1 the problem with the RI defined in [11] is clear since the RI becomes negative before the amplitude of the terrain reaches the ground clearance of the vehicle. With the newly proposed RI from Eq. (1) it can be seen that the RI starts at zero and becomes an ever increasingly positive number as terrain roughness increases. This behavior is more intuitive than the method from [11], therefore it is desirable to use the RI from Eq. (1).

## 3     Velocity Models

In the work presented in [11] a method for determining allowable velocity from a calculated RI and a known allowable excitation force was developed. In this model the calculations depend on an estimated maximum terrain frequency. As a result complications can arise in situations where the maximum terrain frequency exceeds the estimated maximum frequency. This can lead to situations where the predicted maximum velocity causes damage to the vehicle. In this paper three new methods are developed to prevent these problems, the first being the new Instantaneous Elevation Change Model, and the last two being variations on the new Sinusoidal Base Excitation Model.

### 3.1    Instantaneous Elevation Change Model

In the Instantaneous Elevation Change Model (IECM) the worst case scenario of a step change in the terrain elevation is considered (Fig. 2). In this scenario the maximum potential force that could be exerted on a vehicle by terrain of any given roughness can be calculated to an arbitrary statistical confidence to be selected by the designer. From Fig. 2 the following expressions for traversal distance ($X_w$), traversal time ($\Delta t$), and average vertical velocity ($\dot{e}$) can be derived (Eq. (2),(3),(4)).

$$X_w = \sqrt{2er - e^2} \qquad (2)$$

$$\Delta t = \frac{X_w}{V_x} = \frac{\sqrt{2er - e^2}}{V_x} \qquad (3)$$

$$\dot{e} = \frac{e}{\Delta t} = \frac{eV_x}{X_w} = \frac{eV_x}{\sqrt{2er - e^2}} \qquad (4)$$

The value for $X_w$ is saturated when $r = e$, therefore if $e \geq r$ then $X_w = r$.

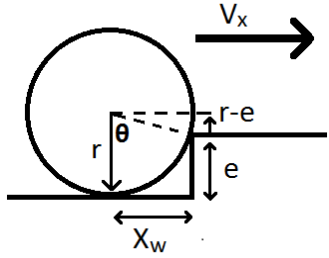
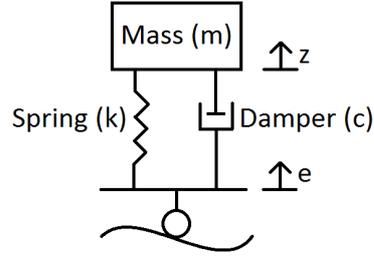

**Fig. 2.** IECM Model             **Fig. 3.** Simplified Quarter Car Model

Using this IECM model the excitation force acting on the vehicle can be derived from a simplified quarter car model (Fig. 3). The derived expression is shown in Eq. (5) where the constants $\omega_n$, $\zeta$, and $m$ represent the natural frequency, damping ratio, and mass of the quarter car model respectively.

$$\hat{F} = \frac{F}{m \cdot \omega_n} = \frac{1}{\omega_n^2}\ddot{z} + \frac{2\xi}{\omega_n}\dot{z} + z = \frac{2\xi}{\omega_n}\dot{e} + e \qquad (5)$$

Combining Eq. (4) and Eq. (5) an expression relating excitation force to velocity is derived:

$$\hat{F} = \left(\frac{2\xi V_x}{\omega_n X_w} + 1\right) e \quad (6)$$

By rearranging Eq. (6) for $V_x$ and substituting $e_{max}$ for $e$, Eq. (7) is obtained.

$$V_x = (\hat{F} - e_{max})\frac{\omega_n X_w}{2\zeta e_{max}} = (\hat{F} - e_{max})\frac{\omega_n \sqrt{2e_{max}r - e_{max}^2}}{2\zeta e_{max}} \quad (7)$$

The expression shown in Eq. (8) for $e_{max}$ is obtained by assuming that the values for $e$ obtained from a laser scanner follow a normal distribution with a mean of 0. With this assumption the cumulative distribution function is used to calculate the probability $(P_e)$ that $e_{max} \leq$ z-score of $e_{max}$. The probability function $P_e$ is then used to obtain Eq. (8).

$$e_{max} = \sqrt{2}\, erf^{-1}(2P_e - 1)(RI)\, h \quad (8)$$

Using Eq. (7) and (8) the allowable traversal velocity can be calculated to an arbitrary confidence, as defined by the designer, through assigning a value for $P_e$ and substituting in an appropriate value for the maximum allowable excitation force $F_{max} = F$ as well as the vehicle and suspension properties.

### 3.2   Sinusoidal Base Excitation Model

With the new IECM the issue from [11] of estimated maximum terrain frequency being exceeded has been avoided. However, since vertical velocity of the terrain is calculated as the average of the vertical velocity during step traversal in IECM, there is the danger that the peak vertical velocity during step traversal could cause damage to the vehicle. To reduce this issue IECM is transformed into a Sinusoidal Base Excitation Model (SBEM) using the traversal time as the quarter period of the function (Fig. 4).

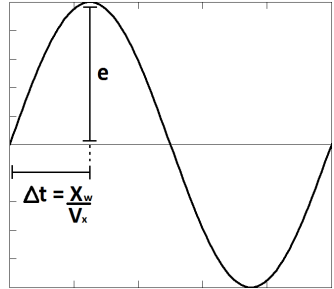

**Fig. 4.** Sinusoidal Base Excitation Model

With the same procedure used to derive Eq. (5), the equation for excitation force was obtained (Eq. (9)). The constants $c$ and $k$ represent the damping constant, and spring constant of the quarter car model respectively.

$$F(t) = m\ddot{z} + c\dot{z} + kz = ke\sin\left(\frac{\pi}{2\Delta t}t\right) + \frac{ce\pi}{2\Delta t}\cos\left(\frac{\pi}{2\Delta t}t\right) \tag{9}$$

Using this new model two new equations relating allowable velocity to the RI are derived in SBEM: Using Excitation Force (SBEM:EF), and SBEM: Using Transmitted Force (SBEM:TF).

**SBEM: Using Excitation Force.**
The SBEM:EF method uses excitation force to define the maximum force that can be exerted on the vehicle. To determine the maximum allowable speed of the vehicle it is important to predict the maximum excitation force during obstacle traversal. To do this the derivative of Eq. (9) is set equal to zero $(\dot{F}(t) = 0)$ and then the equation can be rearranged for time. As shown in Eq. (10) this enables the calculation of the time at which the maximum excitation force occurs $(t_{max})$ with respect to the start of the obstacle traversal.

$$t_{max} = \frac{2X_w}{\pi V_x}\tan^{-1}\left(\frac{2kX_w}{\pi cV_x}\right) \tag{10}$$

Substituting Eq. (10) into Eq. (9) and using Eq. (8) for $e$, an expression representing the maximum excitation force is obtained (Eq. (11)).

$$F_{max} = \frac{\frac{2k^2 e_{max} X_w}{\pi c V_x} + \frac{\pi c e_{max} V_x}{2X_w}}{\sqrt{1 + \frac{4k^2 X_w^2}{\pi^2 c^2 V_x^2}}} \tag{11}$$

As a result the maximum allowable velocity for the vehicle can now be obtained from Eq. (11) with a numerical method using the appropriate value for $F_{max}$ as defined by the designer. In this paper Newton's Method was used to solve for the allowable velocity $V_x$.

**SBEM: Using Transmitted Force.**
In contrast to SBEM:EF, SBEM:TF uses transmitted force to define the maximum force that can be exerted on the vehicle. A procedure for determining transmitted forces in general cases can be found in Section 2.4 of [12]. Adapting this procedure and using $\omega_b$ as defined in Eq. (12), an expression for the maximum transmitted force problem in this paper is obtained in Eq. (13).

$$\omega_b = \frac{\pi V_x}{2X_w} \tag{12}$$

$$F_{t_{max}} = \frac{ke_{max}\pi^2 V_x^2}{4X_w^2 \omega_n^2}\left[\frac{1+\left(\frac{\pi\zeta V_x}{X_w \omega_n}\right)^2}{\left(1-\left(\frac{\pi V_x}{2X_w \omega_n}\right)^2\right)^2 + \left(\frac{\pi\zeta V_x}{X_w \omega_n}\right)^2}\right]^{1/2} \tag{13}$$

As with the excitation force method, the maximum allowable velocity $V_x$ is solved with a numerical method in Eq. (13) using the appropriate value for $F_{t\,max}$ as defined by the designer. As stated previously, Newton's Method was used in this paper.

## 4    Results

In this section the three force prediction methods for determining allowable maximum velocities derived in this paper are compared to each other (Eq. (7), Eq. (11), Eq. (13)) as well as to the method derived in [11]. The methods being compared are IECM, SBEM:EF, SBEM:TF, and Prev:BEM (Base Excitation Model from [11]).

To compare these methods vehicle properties for a typical large All-Terrain Vehicle were assumed to allow for simulation. The vehicle properties can be seen in Table 1. The terrain was simulated by one hundred different 2D sinusoidal profiles with properties also listed in Table 1. The sinusoidal amplitude of the terrain was set at one hundred different values evenly spaced between 0 and 0.1m. The sinusoidal profile was used to represent a typical 2D uniformly oscillating terrain profile.

Table 1. Vehicle/Terrain Properties

| Quarter Mass $m$ | Ground Clearance $h$ | Wheel Radius $r$ | Suspension Natural Frequency $\omega_n$ | Suspension Damping Ratio $\zeta$ |
|---|---|---|---|---|
| 100 kg | 0.1 m | 0.1 m | 3 rad/s | 0.5 |

| Confidence in $e_{max}$ $P_e$ | Maximum Allowable Excitation Force $F_{max}$ | Maximum Allowable Transmitted Force $F_{t_{max}}$ | Terrain Distance Frequency $f$ | Terrain Amplitude |
|---|---|---|---|---|
| 0.999999 | 5000 N | 5000 N | 1.5 Hz | -0.1 to 0.1 m |

Using the vehicle properties defined in Table 1 the methods were compared graphically. As previously stated, one hundred unique amplitudes were used for the sinusoidal terrain profiles; therefore one hundred unique RI values were used for plotting. In Fig. 5 the calculated allowable velocities of the four methods are plotted against the RI of the terrain (calculated from the sinusoidal terrain profiles). In addition, the maximum velocity of the vehicle was considered to be 20 $m/s$ (consistent with a typical large All-Terrain Vehicle).

From Fig. 5 it can be seen that, as expected, the method derived in [11] produces the most aggressive velocity profile. Also as expected, the SBEM:EF and SBEM:TF models produced the safest velocity profiles. IECM had a more aggressive velocity profile than the two SBEM methods, which is consistent with expectations since it relies on the average vertical velocity during obstacle traversal, which as previously stated could lead to unsafe force predictions.

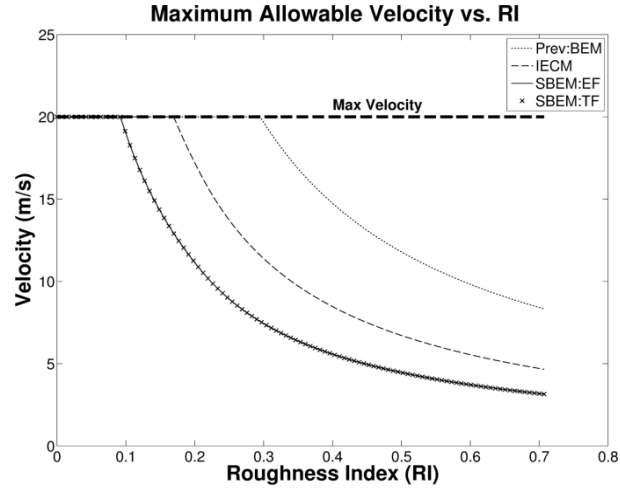

**Fig. 5.** Allowable Velocity vs. RI

It can also be seen that SBEM:EF and SBEM:TF produced almost identical profiles. The transmitted force is very dependent on the frequency ratio $\omega_b/\omega_n$, where for high frequency ratios the transmitted force is generally higher than the excitation force. However, for low frequency ratios the transmitted force can be less than the excitation force [12]. The problem with low frequency ratios is that in this area resonance occurs as $\omega_b/\omega_n \to 1$ ; therefore these ratios are often avoided. In this case the transmitted and excitation force are approximately equal, which is generally the best case scenario when avoiding resonance by using high frequency ratios.

To highlight the differences between SBEM:EF and SBEM:TF the vehicle properties in Table 1 have been manipulated to induce resonance where $\omega_n = 6\ rad/s$, $\zeta = 0.1$, and $F_{max} = F_{t\ max} = 1000\ N$. The graphs comparing these methods in the resonance case are shown in Fig. 6.

With resonance it can be seen that SBEM:TF deviates significantly from SBEM:EF, especially starting at $RI \approx 0.45$. This RI value is where resonance reaches full effect due to the nature of the sinusoidal terrain. It is easy to see that SBEM:TF is a much safer method than all other presented methods since it is very possible for transmitted forces to exceed the excitation force.

9      G.N. Wilson et al.

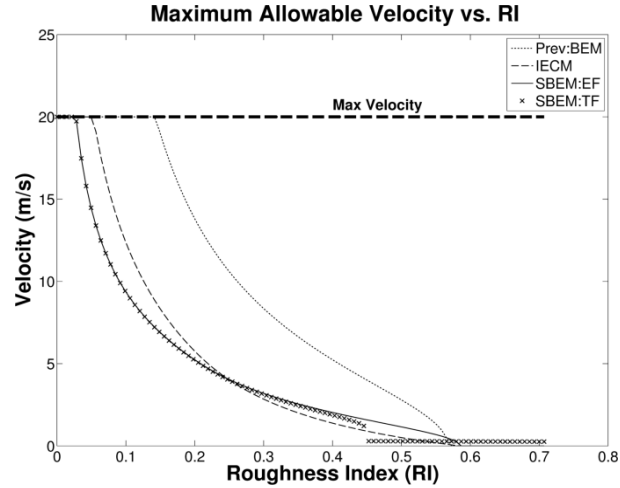

**Fig. 6.** Allowable Velocity vs. RI – Resonance Case

## 5     Conclusion

This paper presents three novel force prediction models for calculating allowable traversal velocity based on a 3D point cloud obtained from proprioceptive sensors. To achieve this a Roughness Index (RI) was used to create a quantitative measure of the upcoming terrain. With this RI three methods for predicting force were derived from the Instantaneous Elevation Change Mode (IECM) and the Sinusoidal Base Excitation Model (SBEM) developed in this paper. The SBEM force prediction was split into two methods, one which predicted the excitation force SBEM:EF, and one which predicted transmitted force SBEM:TF.

The three new methods were compared to each other as well as to the force prediction method developed in [11]. The comparison was done using sets of sinusoidal terrain with different amplitudes. In the comparison it was found that as expected the method developed in [11] produced very aggressive allowable velocities which could cause damage to the vehicle during traversal. The IECM produced the second most aggressive profile since the vertical velocity of the terrain was calculated as an average over the traversal of an obstacle. The SBEM methods produced almost identical results for the particular dynamics selected when resonance was not in effect, both predicting the safest velocity profile. When resonance is being avoided the frequency ratio $\omega_b/\omega_n$ should be high; therefore it is expected that the transmitted force will be at least slightly greater than the excitation force. In the case of resonance it was seen that SBEM:TF produced much different results than SBEM:EF. When resonance frequency was encountered SBEM:TF demonstrated its ability to reduce the allowable velocity to maintain safe traversal speeds.

From the simulations it is expected that the SBEM:EF and SBEM:TF methods would be the safest techniques to use as it is suspected that both the method from [11] and the ICEM method could produce unsafe allowable velocities. In particular

SBEM:TF seems like the most promising candidate since it accounts for resonance and also predicts forces transmitted to the vehicle frame, which are the most important for avoiding equipment damage.

For future work these methods will be tested experimentally on a vehicle platform in various rough terrains to verify simulated expectations. Also since suspension parameters may not always be available an adaptive online approach will be developed to eliminate the need for known suspension parameters. These techniques for determining allowable velocities will also be integrated into a navigation system that will allow UGVs to operate autonomously at high speeds in unknown rough terrains to accomplish predetermined navigation goals.